\documentclass[a4paper,11pt]{article}
\usepackage[round]{natbib}
\usepackage{colortbl}

\pdfoutput=1
\usepackage[plainpages=false]{hyperref}

\newcommand{\xbf}{\mathbf{x}}
\newcommand{\Xbf}{\mathbf{X}}
\newcommand{\hbf}{\mathbf{h}}
\newcommand{\Hbf}{\mathbf{H}}
\newcommand{\ybf}{\mathbf{y}}
\newcommand{\dbf}{\mathbf{d}}
\newcommand{\cbf}{\mathbf{c}}
\newcommand{\bbf}{\mathbf{b}}
\newcommand{\ebf}{\mathbf{e}}
\newcommand{\fbf}{\mathbf{g}}
\newcommand{\Fbf}{\mathbf{G}}
\newcommand{\f}{g}
\newcommand{\Wbf}{\mathbf{W}}
\newcommand{\Ubf}{\mathbf{U}}
\newcommand{\Abf}{\mathbf{A}}
\newcommand{\act}{\mathrm{act}}
\newcommand{\cl}{c}
\newcommand{\Cl}{C}
\newcommand{\bagSize}{{|\Xbf|}}
\newcommand{\sumOverTheBag}{\sum_{s}}
\newcommand{\sumOverTheBagLong}{\sum_{s=1}^{\bagSize}}

\newcommand{\propone}{\text{ClassSetRBM}^{\rm XOR}}
\newcommand{\proponediscr}{$\propone$}
\newcommand{\proponehybrid}{\proponediscr Hybrid}
\newcommand{\proponehardmax}{$\text{ClassSetRBM}^{\rm XOR*}$}
\newcommand{\proponehardmaxhybrid}{\proponehardmax Hybrid}

\newcommand{\proptwo}{\text{ClassSetRBM}^{\rm OR}}
\newcommand{\proptwodiscr}{$\proptwo$}
\newcommand{\proptwohybrid}{\proptwodiscr Hybrid}
\newcommand{\proptwohardmax}{$\text{ClassSetRBM}^{\rm OR*}$}
\newcommand{\proptwohardmaxhybrid}{\proptwohardmax Hybrid}
\newcommand{\BF}{\cellcolor[gray]{0.8}\bf}
\newcommand{\graycell}{\cellcolor[gray]{0.8}}

\def\miGraph{_{mi}}%{_{\mathrm{miGraph}}}
\def\miGraphh{_{mi2}}%{_{\mathrm{miGraph2}}}
\def\kRBF{k}%{k_{rbf}}

\def\LearningRate{\lambda}
\def\LearningRateCD{{\LearningRate}}%_{CD}}}

\def\Sigmoid{\mathrm{sigm}}
\def\Softplus{\mathrm{softplus}}
\def\Softminus{\mathrm{softminus}}%{\log \Sigmoid}

\title{Classification of Sets using Restricted Boltzmann Machines} 

\author{ 
  \begin{tabular}{c}
    J\'er\^ome Louradour \\
    A2iA SA\\
    Paris, France\\
    {\it jl@a2ia.com}\\
    ~
  \end{tabular} 
  \begin{tabular}{c}
    Hugo Larochelle  \\ 
    Department of Computer Science\\
    University of Toronto \\  
    Toronto, Canada\\
    {\it larocheh@cs.toronto.edu}
  \end{tabular}
}
\date{}

\usepackage{tabularx}
\usepackage[pdftex]{graphicx}
\usepackage{color}
\usepackage{amsmath}
\usepackage{amsfonts}
\usepackage{bbold}

\begin{document} 
 
\maketitle 
 
\begin{abstract} 
  We consider the problem of classification when inputs correspond to
  sets of vectors. This setting occurs in many problems such as the
  classification of pieces of mail containing several pages, of web sites
  with several sections or of images that have been pre-segmented into
  smaller regions. We propose generalizations of the restricted
  Boltzmann machine (RBM) that are appropriate in this context and
  explore how to incorporate different assumptions about the
  relationship between the input sets and the target class within the
  RBM. In experiments on standard multiple-instance learning datasets,
  we demonstrate the competitiveness of approaches based on RBMs and
  apply the proposed variants to the problem of incoming mail
  classification.
\end{abstract} 
 
\section{Introduction}

The vast majority of machine learning algorithms are developed in the
context where each input can be assumed to take the form of a
fixed-size vector $\xbf$. In some applications however, such an
assumption does not hold and inputs cannot easily be processed into this
form. In this paper, we consider one such setting where inputs consist
in an unordered and variable-length set of vectors $\Xbf =
\{\xbf^{(1)},\dots,\xbf^{(\bagSize)}\}$.
For instance, $\Xbf$ could be the set of text
documents $\xbf^{(s)}$ found in some incoming piece of mail, where each
document is represented as a bag of words. In this particular example,
a simple approach to converting the set $\Xbf$ into a single vector
$\xbf$ would consist in computing the global bag of word
representation of all documents in $\Xbf$, as if all documents had
been concatenated into a single one. This would however correspond to
throwing away all the information about the structure of the incoming
mail, which could be useful to solve the task at hand.  This problem setting
is not specific to text data either: $\Xbf$ could correspond to a
collection of images or to a single image that has been pre-segmented,
and some recognition tasks in computer vision have previously been
formulated in terms of classification of
sets~\citep{Kondor2003,Wallraven2003}.
Another example is
text-independent speaker recognition~\citep{Reynolds1995}, where inputs
are sequences of acoustic vectors but for which the order is not relevant:
relevant short-term dynamics
are taken into account in the vector features themselves (e.g.\ spectral
coefficients and their derivatives) and long-term dynamics are not
useful for classification (the succession of
these features is informative of the speech content, not the
speaker identity).

A popular approach to classifying sets has been that of 
multiple-instance learning (MIL).
In MIL, binary
classification of sets of vectors is performed by assuming that a set
belongs to the positive class if at least one element of the set
belongs to that positive class. Otherwise the set belongs to the
negative class (i.e.\ all elements of the set are from the negative
class). This problem was originally motivated in the context of drug
activity prediction~\citep{Dietterich1997}, where a drug molecule can take several shapes but
only some of them might allow the molecule to bind with some
given protein associated with a disease. A drug molecule can then be
represented as the set of its potential shapes and this set will have
a positive label only if at least one of its shapes allow binding.

The MIL approach makes the implicit assumption that the presence of just
a single positive example is sufficient to recognize the whole set as
positive. However, this assumption is not always appropriate. 
For instance, each vector could only provide {\it partial} 
class information, such that the observation of only a single
informative vector is not enough to label the whole set as positive. 

In this paper, we describe extensions of the restricted Boltzmann
machine that perform multiclass classification of sets and does
not assume that sufficient discriminative information is present in a
single element of the set. By learning a latent representation of its input,
these extensions can deal with cases where only partial evidence of
class membership is present in only a few set vectors.  We report
competitive results on some common MIL datasets and present an
application of these models to a mail classification problem.

\section{Classification with Restricted Boltzmann Machines}
\label{sec:class_rbm}

In this work, we build on a specific restricted Boltzmann machine
(RBM) that can be used to perform
classification~\citep{Larochelle2008,Tieleman2008}. We will refer to
this RBM as a classification RBM (ClassRBM).

The ClassRBM is an energy-based probabilistic model where a
layer $\hbf$ of $H$ binary hidden units are used to model the joint distribution
of a vector of $D$ inputs $\xbf$ and a target vector $\ybf$ of size
$\Cl$.  The target $\ybf$ corresponds to a class label and takes the
``one out of $\Cl$'' representation, meaning that if $\xbf$ belongs
to class $\cl$, then $\ybf = \ebf_{\cl}$ where $\ebf_{\cl}$ is a
vector with all values set to 0 except at position $\cl$, which is set
to 1. For simplicity, we will also assume that $\xbf$ is a binary
vector, though generalizations to other types of vectors are
possible~\citep{Welling2005}.

\begin{figure*}
\begin{center}
\includegraphics[width=0.4\linewidth]{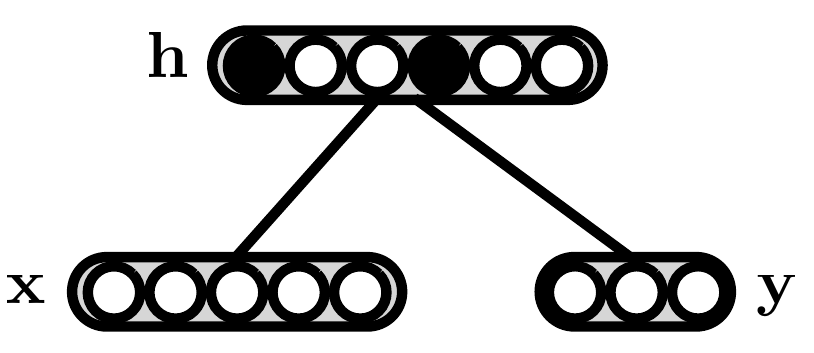}
\end{center}
\caption{Illustration of a standard ClassRBM.
  An example of activations for the hidden layer
  units is given (black means hidden unit is equal to 1).}
\label{fig:classrbm}
\end{figure*}

Using the energy function
\begin{equation*}
E( \xbf, \ybf, \hbf )= - \dbf^\top \ybf - \cbf^\top \hbf  - \bbf^\top \xbf - \hbf^\top \Wbf \xbf   - \hbf^\top \Ubf \ybf,
\end{equation*}
the probability for some configuration of $\xbf$, $\ybf$ and $\hbf$ is defined as
\begin{equation}
\label{eq:energy_based_proba}
p( \xbf, \ybf, \hbf  ) = \exp( - E( \xbf, \ybf, \hbf ) ) / Z
\end{equation}
where $Z$ is a normalizing constant that ensures $p( \xbf, \ybf, \hbf
)$ defines a valid distribution. Figure~\ref{fig:classrbm} shows
an illustration of a ClassRBM. 

Though $Z$ (and hence
$p(\xbf,\ybf,\hbf)$) is usually intractable to compute, the following
conditional distributions of the model are themselves tractable:
\begin{eqnarray*}
p(\hbf | \xbf, \ybf) & = &\prod_{j=1}^H p( h_j | \xbf, \ybf)\\
p(h_j=1 | \xbf, \ybf) & = & \Sigmoid(c_j + \Wbf_{j\cdot} \xbf + \Ubf_{j\cdot}\ybf)\\
p(\xbf,\ybf | \hbf) & = & p(\ybf | \hbf) \prod_{i=1}^D p( x_i | \hbf)\\
p( x_i = 1| \hbf) & = & \Sigmoid(b_i + \hbf^\top \Wbf_{\cdot i})\\
p( \ybf = \ebf_{\cl} | \hbf ) & = & \frac{\exp(\dbf^\top \ybf + \hbf^\top \Ubf \ybf)}{\sum_{\cl'=1}^{\Cl}\exp(\dbf^\top \ebf_{\cl'} + \hbf^\top \Ubf \ebf_{\cl'})}
\end{eqnarray*}
where $\Sigmoid(v) = 1 / (1+\exp(-v))$ and 
we use the notation $\Abf_{j\cdot}$ to refer to the
$j^{\mathrm{th}}$ row of matrix $\Abf$ and $\Abf_{\cdot i}$ for it's $i^{\mathrm{th}}$
column.

Given~(\ref{eq:energy_based_proba}), it can also be shown that the
posterior class probability distribution given some input $\xbf$ 
has the closed form
\begin{eqnarray}
p( \ybf = \ebf_{\cl} | \xbf ) & = & \sum_{\hbf} p(\ybf = \ebf_{\cl}, \hbf | \xbf ) \nonumber\\
 & = & \frac{\exp(-F(\xbf,\ybf))}{\sum_{\cl'=1\ldots \Cl}\exp(-F(\xbf,\ebf_{\cl'}))}\nonumber%\label{eq:rbm_posterior_proba}
\end{eqnarray}
where $F(\xbf,\ybf)$ is referred to as the free-energy
\begin{equation*}
F(\xbf,\ybf) = -\dbf^\top \ybf - \sum_{j=1}^H \Softplus\left( c_j + \Wbf_{j,\cdot} \xbf + \Ubf_{j,\cdot}\ybf \right)
\end{equation*}
with $\Softplus(v)  = \log(1+\exp(v))$. 

In order to train the ClassRBM, different strategies can be followed.
A first option is to train it discriminatively, by minimizing the
average negative conditional log-likelihood $-\log p(\ybf_t | \xbf_t)$
of the parameters for the available training data
$\{\xbf_t,\ybf_t\}$. This can be achieved by simple stochastic
gradient descent.

A second option is to train the ClassRBM generatively, by minimizing
the negative joint log-likelihood $-\log p(\xbf_t,
\ybf_t)$. Unfortunately, the necessary gradients
cannot be computed exactly.
The Contrastive
Divergence (CD) algorithm~\citep{Hinton2006} however provides a useful 
approximation
\begin{align*}
\frac{\partial -\log p(\xbf_t,\ybf_t)}{\partial \theta}  \approx & ~ {\rm E}_{\hbf | \xbf_t, \ybf_t}\Big[\frac{\partial E(\xbf_t,\ybf_t,\hbf)}{\partial \theta}\Big] \nonumber \\
  & - {\rm E}_{\hbf | \widetilde{\xbf}_t, \widetilde{\ybf}_t}\Big[\frac{\partial E(\widetilde{\xbf}_t,\widetilde{\ybf}_t,\hbf)}{\partial \theta}\Big]
\end{align*}
where $\theta$ is any parameter of the ClassRBM
and
where $\widetilde{\xbf}_t$ and $\widetilde{\ybf}_t$ is the result of a
one-step Gibbs sampling chain, initialized at the training example
$\xbf_t$ and $\ybf_t$. 
Noting $\widehat{\hbf}_t = \Sigmoid(\cbf + \Wbf \xbf_t + \Ubf\ybf_t)$ and
$\widetilde{\hbf}_t = \Sigmoid(\cbf + \Wbf \widetilde{\xbf}_t + \Ubf\widetilde{\ybf}_t)$, 
we get the following stochastic gradient update from CD:
\begin{equation*}\begin{array}{lclcl}
\bbf & \leftarrow & \bbf &+& \LearningRateCD~(\xbf_t - \widetilde{\xbf}_t)\\
\cbf & \leftarrow & \cbf &+& \LearningRateCD~(\widehat{\hbf}_t - \widetilde{\hbf}_t)\\
\dbf & \leftarrow & \dbf &+& \LearningRateCD~(\ybf_t - \widetilde{\ybf}_t)\\
\Wbf & \leftarrow & \Wbf &+& \LearningRateCD~(\widehat{\hbf}_t \xbf_t^\top  - \widetilde{\hbf}_t \widetilde{\xbf}_t^\top)\\
\Ubf & \leftarrow & \Ubf &+& \LearningRateCD~(\widehat{\hbf}_t \ybf_t^\top  - \widetilde{\hbf}_t \widetilde{\ybf}_t^\top)
\end{array}\end{equation*}
where $\LearningRateCD$ is the stochastic gradient learning rate used for generative training.

As argued by \citet{Larochelle2008}, in some situations, neither
discriminative or generative learning alone are optimal and better
performance can be achieved by using a linear combination of both
objectives. This is referred to as hybrid generative/discriminative
learning and corresponds to performing both the discriminative and
generative parameter updates with separate learning rates.

\section{Generalization of the ClassRBM to handle sets (ClassSetRBM)}

Now, we wish to generalize the ClassRBM so that it can model the
distribution of a set $\Xbf = \{\xbf^{(1)},\dots,\xbf^{(\bagSize)}\}$
with target vector $\ybf$.  The simplest approach would be to
connect each vector $\xbf^{(s)}$ to some global hidden layer $\hbf$
with the same connection matrix $\Wbf$. However, this approach is not
appropriate because, not only do we expect sets to have varying sizes,
but also the number of vectors $\xbf^{(s)}$ in $\Xbf$ that actually
contain predictive information about $\ybf$ will also vary. By having
just a single global hidden layer, the activity of hidden units would
thus tend to over-saturate for sets of large size.

To address this issue, we propose two generalizations where each vector
$\xbf^{(s)}$ of a set will be connected to its own ``copy'' of the
hidden layer. The number of hidden layers will then depend on the size
of the input\footnote{The implication of this is that we always
  condition on the size of the input set.}. All hidden layers will be
connected to its corresponding input vector by the same matrix $\Wbf$.

Given this approach, there are still different design choices to
be made as to how these hidden layers should interact with the
target units $\ybf$. We present here two such choices,
which correspond to different assumptions about the nature of the
interaction between the input sets and the target.

\subsection{ClassSetRBM with Mutually Exclusive Hidden Units (XOR)}
\label{ssec:prop1}

If we believe that the vectors in the input set $\Xbf$ all contain
information of a distinct nature, then a hidden feature detected
within one vector $\xbf^{(s)}$ would be expected to be absent
in the other vectors of the set. In this case, the set structure
would convey very useful information about how to perform 
classification. 

\begin{figure*}
\begin{center}
\includegraphics[width=0.45\linewidth]{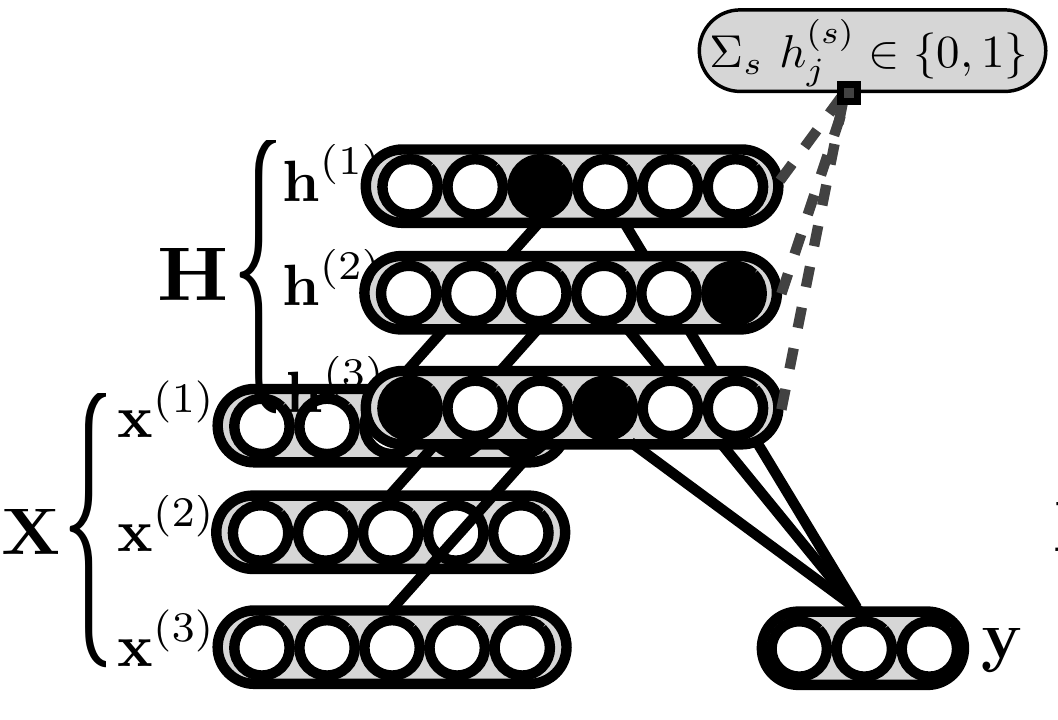}
\end{center}
\caption{Illustration of $\propone$. Dotted lines connect hidden layers whose activity is
  subject to joint constraints. An example of valid activations for the hidden layer
  units is given.}
\label{fig:classsetrbm_xor}
\end{figure*}

To exploit such information, we could impose that the activity of
hidden units be mutually exclusive across the vectors of the
set. Noting $\hbf^{(s)}$ the hidden layer to which $\xbf^{(s)}$ is
connected and $\Hbf = \{\hbf^{(1)},\dots,\hbf^{(\bagSize)}\}$ the set of
hidden layer vectors, this would translate into requiring the
constraint that
\begin{align*}
  \textstyle
  \sumOverTheBagLong
      h_j^{(s)} \in \{0,1\} \qquad \forall j=1\ldots H
\end{align*}
i.e.\ for all hidden unit position $j$, at most one hidden unit $h_j^{(s)}$
should be active across all vectors $\xbf^{(s)}$. With that, we define the
energy function
\begin{align*}
E(\Xbf, \ybf, \Hbf) = -\dbf^\top \ybf
- \sumOverTheBag
      \bbf^\top \xbf^{(s)}
- \sumOverTheBag
      \cbf^\top \hbf^{(s)} 
- \sumOverTheBag
      \Big( {\hbf^{(s)}}^\top \Wbf \xbf^{(s)}  + {\hbf^{(s)}}^\top \Ubf \ybf \Big)\nonumber
\end{align*}
where the target $\ybf$ is connected to all hidden layers through the
same connection matrix $\Ubf$. We will refer to this variant of the ClassRBM
for sets as $\propone$. See Figure~\ref{fig:classsetrbm_xor} 
for an illustration of $\propone$.

While more complicated than the ClassRBM for single vectors, it
can be shown that $\propone$ has simple conditional distributions as
well. The hidden layers conditional distribution becomes
\begin{equation*}\begin{array}{rcl}
p(\Hbf | \Xbf, \ybf) & = & \prod_{j=1}^H p\big( \{h^{(s)}_j\}_{s=1}^{\bagSize} | \Xbf, \ybf \big)\\
p(h^{(s)}_j = 1 | \Xbf, \ybf) & = & \frac{\exp\left(\act_j(\xbf^{(s)},\ybf)\right)}{1 + \sum_{s'=1}^{\bagSize} \exp\left(\act_j(\xbf^{(s')},\ybf)\right)}\\
p(h^{(\cdot)}_j = 0 | \Xbf, \ybf) & = & \frac{1}{1 + \sum_{s'=1}^{\bagSize} \exp\left(\act_j(\xbf^{(s')},\ybf)\right)}
\end{array}\end{equation*}
where $\act_j(\xbf^{(s)},\ybf) = c_j + \Wbf_{j\cdot} \xbf^{(s)}
+ \Ubf_{j\cdot}\ybf$ and the statement $h^{(\cdot)}_j = 0$ is a
shorthand for $h^{(s)}_j = 0~\forall s=1,\dots,\bagSize$.
The input and target vectors' distribution are
\begin{equation*}\begin{array}{rcl}
p(\Xbf,\ybf | \Hbf) & = & p(\ybf | \Hbf) \prod_{s=1}^{\bagSize} \prod_{i=1}^D p( x^{(s)}_i | \hbf^{(s)})\\
p( x^{(s)}_i = 1| \hbf^{(s)}) & = & \Sigmoid(b_i + {\hbf^{(s)}}^\top \Wbf_{\cdot i})\\
p( \ybf = \ebf_{\cl} | \Hbf ) & = & \frac{\exp(\dbf^\top \ybf +
  \sum_{s=1}^{|X|} {\hbf^{(s)}}^\top \Ubf
  \ybf)}{\sum_{\cl'=1}^{\Cl}\exp(\dbf^\top \ebf_{\cl'} +
  \sum_{s=1}^{|X|} {\hbf^{(s)}}^\top \Ubf \ebf_{\cl'})} .
\end{array}\end{equation*}

These conditional distributions are simple enough that
Gibbs sampling can be performed, by sampling each
element of $\Hbf$ given $\Xbf$ and $\ybf$, and then
sampling new values for the vectors in $\Xbf$ and
for $\ybf$.

The target posterior $p(\ybf | \Xbf)$ can also be
computed efficiently. It can be shown that it has the
following form
\begin{align}
p( \ybf = \ebf_{\cl} | \Xbf )  =  \frac{\exp(-F^{\rm XOR}(\Xbf,\ybf))}{\sum_{\cl'=1\ldots \Cl}\exp(-F^{\rm XOR}(\Xbf,\ebf_{\cl'}))}\nonumber%\label{eq:rbm_posterior_proba_propone}
\end{align}
where the free-energy $F^{\rm XOR}(\Xbf,\ybf)$ is now 
\begin{equation*}
F^{\rm XOR}(\Xbf,\ybf) = -\dbf^\top \ybf  - \sum_{j=1}^H \Softplus\left( {\rm softmax}_j(\Xbf) +\Ubf_{j\cdot}\ybf \right) 
\end{equation*}
with ${\rm softmax}_j(\Xbf) = \log( \sum_{s=1}^{\bagSize} \exp(c_j +
\Wbf_{j\cdot} \xbf^{(s)}))$ can be seen as a soft version of the
max function of $c_j + \Wbf_{j\cdot} \xbf^{(s)}$ over the set of input
vectors.

As before, given a training set of pairs $\{\Xbf_t,\ybf_t\}$, it is
possible to train $\propone$ discriminatively and generatively. The
discriminative gradients are easily computed using the chain rule. CD
approximations for the generative learning updates can also be
obtained, since Gibbs sampling can be performed.

We note $\widetilde{\Xbf}_t =
\{\widetilde{\xbf}_t^{(1)},\dots,\widetilde{\xbf}_t^{(|\Xbf_t|)}\}$ and $\widetilde{\ybf}_t$
as the result of one step of Gibbs sampling initialized at the
training example pair $(\Xbf_t,\ybf_t)$. Similarly to the standard
ClassRBM case, we also note $\widehat{\hbf}^{(s)}_t$ and
$\widetilde{\hbf}^{(s)}_t$ as the vector containing the conditional
probabilities of the hidden units being equal to 1, conditioned on
$(\Xbf_t,\ybf_t)$ and $(\widetilde{\Xbf}_t,\widetilde{\ybf}_t)$
respectively.  Then, the Contrastive Divergence (CD) learning updates
are computed as follows:
\def\sumOverBag{\sum_{s}}
\begin{equation*}\begin{array}{lclcl} 
\bbf & \leftarrow & \bbf &+& \LearningRateCD~\sumOverBag \big( \xbf^{(s)}_t - \widetilde{\xbf}^{(s)}_t\big)\\
\cbf & \leftarrow & \cbf &+& \LearningRateCD~\sumOverBag \big(\widehat{\hbf}^{(s)}_t -  \widetilde{\hbf}^{(s)}_t\big)\\
\dbf & \leftarrow & \dbf &+& \LearningRateCD~\big(\ybf_t - \widetilde{\ybf}_t\big)\\
\Wbf & \leftarrow & \Wbf &+& \LearningRateCD~\sumOverBag \big(\widehat{\hbf}^{(s)}_t {\xbf_t^{(s)}}{}^\top  - \widetilde{\hbf}^{(s)}_t \widetilde{\xbf}^{(s)}_t{}^\top \big)\\
\Ubf & \leftarrow & \Ubf &+& \LearningRateCD~\sumOverBag \big(\widehat{\hbf}_t^{(s)} \ybf_t^\top  - \widetilde{\hbf}_t^{(s)} \widetilde{\ybf}_t^\top\big).
\end{array}\end{equation*} 

\subsection{ClassSetRBM for Sets with Redundant Evidence (OR)}
\label{ssec:prop2}

There might be cases where the assumption of mutual exclusivity of the
hidden units is too strong. One simple such case would be if
additional copies of vectors were inserted in the set. More generally,
it could be that the same useful hidden feature is
present in several vectors within the input set. In this case, the
actual number of vectors containing this evidence is not
relevant, only the presence of that evidence in at least one set
element is. It might then be desirable to have a model that is more
robust to such situations.

\begin{figure*}
\begin{center}
\includegraphics[width=0.65\linewidth]{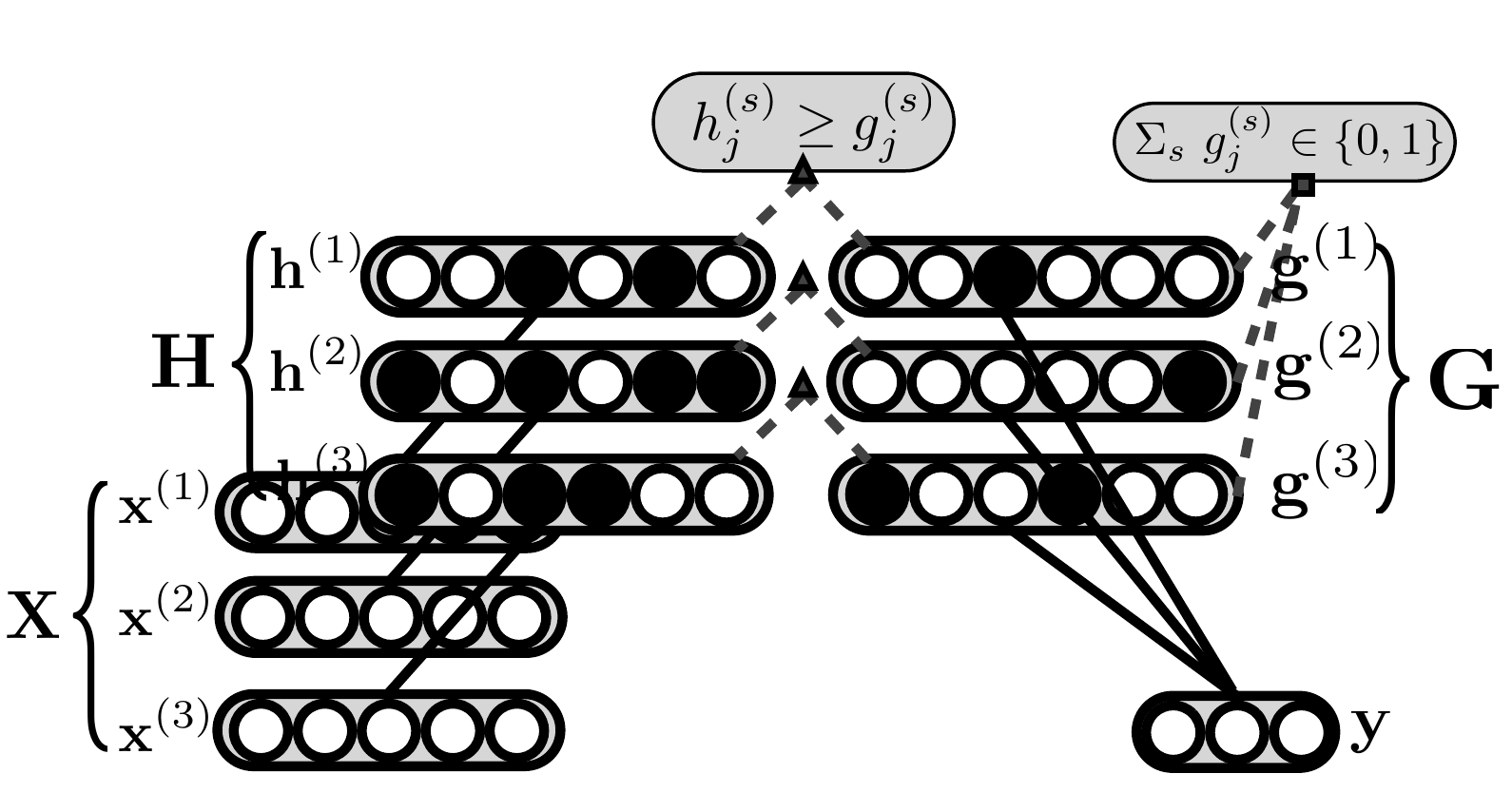}
\end{center}
\caption{Illustration of $\proptwo$. Compared to Figure~\ref{fig:classsetrbm_xor}, 
  this RBM has a set of pairs of hidden layers (one for the inputs, one for the target),
  with additional constraints within pairs. The multual exclusivity constraint only
  applies to the target hidden layers. An example of valid activations for the hidden layer
  units is also given.}
\label{fig:classsetrbm_or}
\end{figure*}

To achieve this, we must somehow remove mutual exclusivity over
the vectors $\xbf^{(s)}$ but maintain it for the connections with the
target $\ybf$. This can be accomplished by having additional hidden
layer ``copies'' $\Fbf = \{\fbf^{(1)},\dots,\fbf^{(\bagSize)}\}$
connected only to $\ybf$ and removing the direct connections of $\Hbf$ to
$\ybf$, yielding the new energy function
\begin{align*}
\textstyle
E(\Xbf, \ybf, \Hbf,\Fbf) = -\dbf^\top \ybf
   - \sumOverTheBag
      \bbf^\top \xbf^{(s)}
   - \sumOverTheBag
      \cbf^\top \hbf^{(s)} 
\textstyle
   - \sumOverTheBag
      \Big( {\hbf^{(s)}}^\top \Wbf \xbf^{(s)}  + {\fbf^{(s)}}^\top \Ubf \ybf \Big)~.\nonumber
\end{align*}
Then, dependencies between $\Xbf$ and $\ybf$ are modeled by the hidden
units through the imposition of
the following constraints in the activities in $\Hbf$ and $\Fbf$:
\begin{eqnarray*}
\textstyle
\sumOverTheBagLong
\f_j^{(s)} \in \{0,1\} &&\forall j=1\dots H\\ h^{(s)}_j \geq \f^{(s)}_j  &&\forall j=1\dots H,~\forall s=1,\dots,\bagSize~.
\end{eqnarray*}
Hence, for a target hidden unit $\f^{(s)}_j$ to be active,
{\it at least} one input hidden units $h^{(s)}_j$ will need
to be active. We call this second model 
$\proptwo$. Also see Figure~\ref{fig:classsetrbm_or} for an 
illustration of $\proptwo$.

It can be shown that $\proptwo$ has the following conditionals over
$\Fbf$ and $\Hbf$:
\begin{equation*}\begin{array}{rcl}
p(\Fbf | \Xbf, \ybf) & = & \prod_{j=1}^H p( \{\f^{(s)}_j\}_{s=1}^{\bagSize} | \Xbf, \ybf)\\
p(\f^{(s)}_j = 1 | \Xbf, \ybf) & = & \frac{\exp\left(\act_j(\xbf^{(s)},\ybf)\right)}{1 + \sum_{s'=1}^{\bagSize} \exp\left(\act_j(\xbf^{(s')},\ybf)\right)}\\
p(\f^{(\cdot)}_j = 0 | \Xbf, \ybf) & = & \frac{1}{1 + \sum_{s'=1}^{\bagSize} \exp\left(\act_j(\xbf^{(s')},\ybf)\right)}
\end{array}\end{equation*}
\begin{equation*}\begin{array}{rcl}
p(\Hbf | \Fbf, \Xbf, \ybf) & = & \prod_{s=1}^{\bagSize} \prod_{j=1}^H p( h^{(s)}_j | \f^{(s)}_j, \xbf^{(s)})\\
p( h^{(s)}_j = 1 | \f^{(s)}_j, \xbf^{(s)}) & = & \left\{ \begin{array}{l} 1, \text{ if }\f^{(s)}_j = 1\\ \Sigmoid(c_j + \Wbf_{j\cdot} \xbf^{(s)}), \text{ else}\end{array} \right.
\end{array}\end{equation*}
where we now have $\act_j(\xbf^{(s)},\ybf) = \Softminus(c_j + \Wbf_{j\cdot} \xbf^{(s)}) + \Ubf_{j\cdot}\ybf$,
with $\Softminus(a)=a-\Softplus(a)$.
The conditionals for $\Xbf$ and $\ybf$ are the same as in Section~\ref{ssec:prop1}, with the exception that $\ybf$ is conditioned on $\Fbf$, not $\Hbf$. 

The class posterior is also tractable. It is the same as in Section~\ref{ssec:prop1}, but with
a free-energy $F^{\rm OR}(\Xbf,\ybf)$ where the ${\rm softmax}_j(\Xbf)$ function is now
\begin{equation*}
\textstyle
{\rm softmax}_j(\Xbf) = \log\Big( \sumOverTheBagLong \exp( \Softminus(c_j +
\Wbf_{j\cdot} \xbf^{(s)}))\Big).
\end{equation*}

Once again, discriminative and generative learning can be performed. When
performing Gibbs sampling to compute the CD updates, the hidden
layers $\Fbf$ and $\Hbf$ samples are obtained by first sampling from
$p(\Fbf | \Xbf, \ybf)$ and then sampling from $p(\Hbf | \Fbf, \Xbf,
\ybf)$. Again denoting $(\widehat{\hbf}^{(s)}_t,\widehat{\fbf}^{(s)}_t)$ 
and $(\widetilde{\hbf}^{(s)}_t,\widetilde{\fbf}^{(s)}_t)$ 
the vectors of hidden
probabilities from conditioning on 
$(\Xbf_t,\ybf_t)$ and $(\widetilde{\Xbf}_t,\widetilde{\ybf}_t)$
respectively, we obtain the following CD updates:
\begin{equation*}\begin{array}{lclcl} 
\bbf & \leftarrow & \bbf &+& \LearningRateCD~\sumOverBag \big( \xbf^{(s)}_t - \widetilde{\xbf}^{(s)}_t\big)\\
\cbf & \leftarrow & \cbf &+& \LearningRateCD~\sumOverBag \big(\widehat{\hbf}^{(s)}_t -  \widetilde{\hbf}^{(s)}_t\big)\\
\dbf & \leftarrow & \dbf &+& \LearningRateCD~\big(\ybf_t - \widetilde{\ybf}_t\big)\\
\Wbf & \leftarrow & \Wbf &+& \LearningRateCD~\sumOverBag \big(\widehat{\hbf}^{(s)}_t {\xbf_t^{(s)}}{}^\top  - \widetilde{\hbf}^{(s)}_t \widetilde{\xbf}^{(s)}_t{}^\top \big)\\
\Ubf & \leftarrow & \Ubf &+& \LearningRateCD~\sumOverBag \big(\widehat{\fbf}_t^{(s)} \ybf_t^\top  - \widetilde{\fbf}_t^{(s)} \widetilde{\ybf}_t^\top\big)~.
\end{array}\end{equation*}

\subsection{ Variants with ``hard'' max pooling}

In both proposed models, the class posterior $p(\ybf | \Xbf)$ require
that some hidden units be implicitly ``pooled'' by taking a soft version
of the maximum over the input set. Specifically, this is achieved through
the $\log( \sum_{s=1}^{|\Xbf} \exp(\cdot))$ operation in their definitions
of ${\rm softmax}_j(\Xbf)$.

In practice, softmax pooling has the disadvantage that at the beginning 
of training, pooling essentially corresponds to summing the activations 
of all hidden units and does not actually select a single hidden unit. 
This potentially makes it harder for the hidden units to specialize.

Hence, we also consider ``hard'' max variants of $\propone$ and $\proptwo$,
referred to as \proponehardmax and \proptwohardmax respectively, where
the $\log( \sum_{s=1}^{|\Xbf} \exp(\cdot))$ operation is replaced by a
$\max_{s=1,\dots,|\Xbf|} (\cdot)$ operation in their definition of ${\rm softmax}_j(\Xbf)$.
This modification is only applied for computing $p(\ybf | \Xbf)$ and the
discriminative gradients. This can be understood as an estimation of the
true class posterior $p(\ybf | \Xbf)$ by approximating parts of the sum over all the hidden
units with a maximum, and optimizing the conditional log-likelihood of that approximation.

\section{Related Work}

As previously mentioned, the problem of classifying inputs
corresponding to sets is closely related to that of multiple-instance
learning (MIL).
The standard case is binary classification, where a set of training vectors is labeled positive
if and only if at least one instance in the set is positive.
The MIL has been studied to solve problems
such as drug activity detection~\citep{Dietterich1997}
and natural scene categorization.
In the last fifteen years,
several types of approaches have been proposed to address MIL,
such as Learning Axis-Parallel Concepts~\citep{Dietterich1997},
Diverse Density~\citep{Maron1998} and its Expectation-Maximization version~\citep{Zhang2001}.
Extensions of
k-nearest neighbours (Citation kNN in \citet{Wang2000}),
Support Vector Machines (MI-SVM in \citet{Andrews2002}),
decisions trees~\citep{Zucker2001},
perceptrons~\citep{Sabato2010}
and neural networks~\citep{Zhou2002} have also been explored. 
Classification within these approaches mainly consists
of computing the maximum output over the vectors in the set,
and a loss function to optimize is expressed accordingly.
The approach presented in this paper
rather consists in performing the (soft) maximum pooling over the sets
in an intermediate latent representation,
instead of on the output posterior probabilities 
as proposed by~\citep{Zhou2002,Sabato2010}.

Concerning classification of sets in general,
several kernels between sets of vectors have been proposed
to generalize kernel-based classifiers (SVM)
without modifying the standard optimization problem.
There are kernels defined between probabilistic density functions 
estimated on each set of vectors, such as the Fisher kernel~\citep{Jaakkola1998},
Mutual Information kernels~\citep{Seeger2002},
Probability product kernels~\citep{Jebara2004,Lyu2005}
and radial  kernels based on a probabilistic distance 
such as Kullback-Leibler \citep{Kondor2003}.
These approaches have been developed with
some particular families of density functions such as
Gaussian distributions and Gaussian mixture models,
and are not convenient when inputs are high-dimensional or sparse.
There are also kernels based on combinations of kernels between 
inter-sets pairs of instances.
These include several kinds of linear combination of kernels on inter-sets 
pairs of instances~\citep{Louradour2007,Zhou2009}
as well as max kernels~\citep{Wallraven2003}.
The mi-Graph kernels of \citet{Zhou2009} actually achieves some of the best 
results reported on MIL tasks~\citep{Deselaers2010}.

The main disadvantage of such kernel-based SVM approaches is 
that they tend not to scale well with big datasets:
the complexity of optimizing an SVM is quadratic in the number of training samples,
and also the complexity of computing kernels between 
sets is quadratic in the number of vectors per set.

Finally, in the RBM literature, \citet{LeeH2009} also explored the
use of soft (probabilistic) pooling operations in a convolutional
RBM. The two models proposed here can be seen as other pooling-based
RBMs that are appropriate when the inputs are sets.

\section{Experiments}

\newcommand{\SVMmiGraph}{SVM-miGraph}
\newcommand{\SVMmiGraphtwo}{SVM-miGraph2}
\newcommand{\SVMlocal}{SVM-max
}
\newcommand{\inputPoolRBM}{ClassRBM-poolIn}
\newcommand{\outputPoolRBM}{ClassRBM-maxOut}
\newcommand{\outputPoolLogit}{Logit-maxOut}
\newcommand{\outputPoolMLP}{MLP-maxOut}

We present here experiments on standard MIL datasets as well as on the
problem of mail classification, which motivated this work.
To evaluate the proposed RBMs for sets,
we compare them with the following baseline models:
\begin{description}
\item[\inputPoolRBM:]
   In this system, we simply apply ClassRBM described in section~\ref{sec:class_rbm}
   on fixed-size vectors that are generated by pooling all the vectors in each input set
   using the maximum, minimum and average values of the vectors' features over that set.
\item[\outputPoolRBM:]
   This model is an implementation for a ClassRBM of the ideas in \citet{Zhou2002,Sabato2010}.
   The model is trained by gradient descent to predict the target based only on
   the input vector that gives the maximal output response.
   We also apply the same strategy
   with logistic regression (\outputPoolLogit) and a one hidden layer perceptron (\outputPoolMLP).
   Note that these methods 
   are only applicable in the case of binary classification,
   so we only use these baselines for MIL problems\footnote{We tried some variants to generalize to multiclass, but the performance was always poor.}.
\item[\SVMmiGraph:]
   This state-of-the-art SVM model based on a kernel between sets
   gave some of the best results on MIL tasks as reported by~\citet{Deselaers2010}.
   The miGraph kernel~\citep{Zhou2009} is:
$$ K\miGraph\big(\mathbf{X_1},\mathbf{X_2}\big)= \frac{ \sum_{s} \sum_{s'} w_s(\mathbf{X_1}) w_{s'}(\mathbf{X_2}) \kRBF(  \mathbf{x_1}^{(s)},  \mathbf{x_2}^{(s')} ) }{ \sum_{s} w_s(\mathbf{X_1}) \sum_{s'}  w_{s'}(\mathbf{X_2})  } $$
   where the vectorial kernel~$\kRBF$ is the Gaussian kernel 
   and where the weights assigned to a vector within a set is inversely proportional to the number of ``edges'' that can be drawn with the other vectors in the same set:
   \begin{equation*}
   \textstyle
   w_s(\mathbf{X})= 1/
   \sum_{s'} {\mathbb 1}_{\|  \mathbf{x}^{(s)} -  \mathbf{x}^{(s')}  \| < \sigma(\mathbf{X}) }
   \end{equation*}
   given an adaptive distance threshold $\sigma(\mathbf{X})$ defined as the average pairwise distance within the set:
   \begin{equation*}
   \textstyle
   \sigma\miGraph(\mathbf{X})= \frac{T(T-1)}{2} \sum_{s \neq s' } \|  \mathbf{x}^{(s)} -  \mathbf{x}^{(s')}  \|~.
   \end{equation*}
\item[\SVMmiGraphtwo:]
   This system is a variant of \SVMmiGraph~where
   the distance threshold to compute the graphs is the same for all sets\footnote{Personal communication with the authors of \citep{Zhou2009}.}
   $\sigma\miGraphh(\mathbf{X})= \sigma_0$.
   This hyper-parameter is tuned by validation,
   such as the $C$ in the SVM loss and the $\gamma$ of the Gaussian kernel.
\item[\SVMlocal:]  
   Like miGraph kernels, local kernels for sets \citep{Wallraven2003}
   are computed from Gaussian kernel values on all inter-sets pairs of vectors.
   Instead of computing a weighted average,
   only the kernel values that are maximal for each vector are summed.   
\end{description}

In all our experiments,
we perform k-fold cross validation,
and at each fold the model hyper-parameters are optimized on a subset of training inputs (20\%),
not used to train the model.
The reported results are obtained by averaging over all test fold examples.
For all models except the SVMs,
we train by stochastic gradient descent, and the hyper-parameters
are the learning rate and the number of updates (early-stopping).
The number of hidden neurons used was 100 (varying this number had little 
influence on the results). RBMs were either trained discriminatively
only or using the hybrid objective.

\subsection{Experiments on MIL benchmark}

\begin{table*}[t]
\newcommand{\tablesize}{\footnotesize}
\centering
\label{tab:res_mil}
\begin{tabular}{ >{\tablesize}l|>{\tablesize}c>{\tablesize}c>{\tablesize}c>{\tablesize}c>{\tablesize}c|>{\tablesize}c }
\hline
Model                   & Musk1     & Musk2     & Elephant  & Fox       & Tiger     & Average \\
\hline\hline
\proponediscr           & \bf 83.04 &     80.39 &     82.30 &     58.50 & \bf 82.40 &  77.33  \\
\proponehybrid          & \bf 84.57 & \bf 84.12 &     82.80 &     55.70 & \bf 82.10 &  77.86  \\
\proponehardmax         & \bf 83.91 & \bf 84.12 & \BF 87.80 & \bf 60.30 & \bf 82.60 &  \graycell 79.75  \\
\proponehardmaxhybrid   & \bf 83.70 & \bf 81.18 & \bf 86.40 &     58.00 & \BF 83.20 &  78.50  \\
\hline
\proptwodiscr           &     82.61 & \bf 83.73 &     82.70 & \bf 58.60 & \bf 80.90 &  77.71  \\
\proptwohybrid          & \bf 84.35 & \BF 84.71 &     82.60 &     55.50 & \bf 82.70 &  77.97  \\
\proptwohardmax         & \bf 85.65 &     80.39 & \bf 87.10 & \bf 59.70 & \bf 82.60 &  79.09  \\
\proptwohardmaxhybrid   & \bf 85.87 & \bf 81.76 &     85.50 &     56.60 & \bf 82.60 &  78.47  \\
\hline\hline
\inputPoolRBM           &     81.52 & \bf 81.37 &     82.70 & \bf 59.80 &     76.80 &  76.44  \\
\outputPoolRBM          & \bf 83.91 &     80.98 &     81.60 &     57.60 &     75.50 &  75.92  \\
\outputPoolMLP          & \bf 85.65 &     78.82 &     82.00 &     55.40 &     74.40 &  75.25  \\
\outputPoolLogit        &     81.09 &     80.00 &     81.90 & \bf 58.80 &     75.90 &  75.54  \\
\hline
\SVMmiGraph             & \BF 86.74 & \bf 82.35 &     83.80 & \BF 61.50 & \bf 81.20 &  79.12  \\
\SVMmiGraphtwo          & \bf 85.43 & \bf 82.35 &     83.80 & \bf 61.30 & \bf 80.80 &  78.74  \\
\SVMlocal               &     83.48 & \bf 84.51 &     84.60 & \bf 59.70 & \bf 81.70 &  78.80  \\
\hline
\end{tabular}
\caption{
Classification accuracies (\%) on MIL datasets
}
\end{table*}

We start by evaluating our approach
on the public and popular MIL datasets\footnote{\scriptsize\texttt{http://www.cs.columbia.edu/$\sim$andrews/mil/datasets.html}}:
{\it Musk1}, {\it Musk2} (drug activity prediction task)
and
{\it Elephant}, {\it Fox}, {\it Tiger} (image annotation task).
We carried out 5-times repeated 10-fold cross-validation. 
Table~\ref{tab:res_mil}
shows the results of the several proposed variants of ClassSetRBM and of the baseline models.
For each dataset,
the best performance is indicated in a gray cell,
and results in bold are the ones with no significant difference with this best reference,
based on a 95\% two-sided Student's t-test on the classification accuracy differences.
Overall, we see the ClassSetRBMs obtain good results compared to the many baselines.
In particular, the best performing variant, {\proponehardmax}, has the highest average
accuracy over all datasets and is never statistically worse than the best reference.
Most importantly, {\proponehardmax} clearly outperforms {\inputPoolRBM} and {\outputPoolRBM} ,
which confirms the usefulness of having a pooling mechanism at the level of the
hidden layer, as opposed to at the input or output level.
Hybrid training does not clearly improve over purely discriminative training. This might
be explained by the fact that the ClassSetRBMs modeled the input units (scaled
in $[0,1]$) as
binary variables. 
The use of a ``hard'' max pooling however does appear to be quite useful
and almost consistently improves on the softmax variant.

\subsection{Experiments on mail categorization}

The task which motivated this work is image document categorization,
where the documents are pieces of mail that can be considered as sets
of pages.  These pages can be printed or handwritten letters, official papers, 
forms, envelops or white pages.  The main
application is mailroom automation, which is of great interest for
large organizations where the volume of incoming mail can reach tens of
thousands per day and must be processed within a couple of days.
Routing of documents can then be done automatically with a
classifier embedded in the document management system.

Each image of a page is processed by an OCR for printed and handwritten
text, which produces binary input features that correspond to the
presence/absence of a given word in the page.  The vocabulary size is
limited to the 10\,000 most frequently recognized words.  Other
features from image analysis are also appended:
\begin{itemize}
\item Sub-resolution (48~features):
These are average gray-scaled pixel values
on a $6\times8$ regular grid.
\item Document Layout Analysis 
(17~features):
The document image is segmented into zones corresponding to boxes, lines,
printed and handwritten text.
We then compute some geometric statistics on each type of zones.
\item Predefined page class detectors (18~features):
Each detector was designed to detect a common type of pages, such as
bank checks, cursive/printed letters and different kinds of official papers.
The output recognition score is used as a feature.
\end{itemize}

The resulting feature vectors are high-dimensional, sparse, and noisy
to some extent (the word error rate of OCR typically lies between 5\%
and 50\%).  The application of mail classification typically does not
fit well the assumptions made in MIL.  In particular, we expect each
page to provide only partial clues for predicting the set label, such
that considering label assignments at the page level is not natural.  
In other words, while there might be enough informative pages
to confidently identify a set's label, this does not imply that this
label is appropriate for any one of these pages individually. Moreover,
this problem is a multiclass classification problem, while most MIL
algorithms are developed for binary classification and do not always
generalize clearly to the multiclass setting, because of the asymmetric
definition (at least one positive label vs.\ all negative labels) of
MIL.

\begin{figure*}[t]
\begin{center}
\begin{minipage}{0.49\linewidth}
\includegraphics[width=\linewidth]{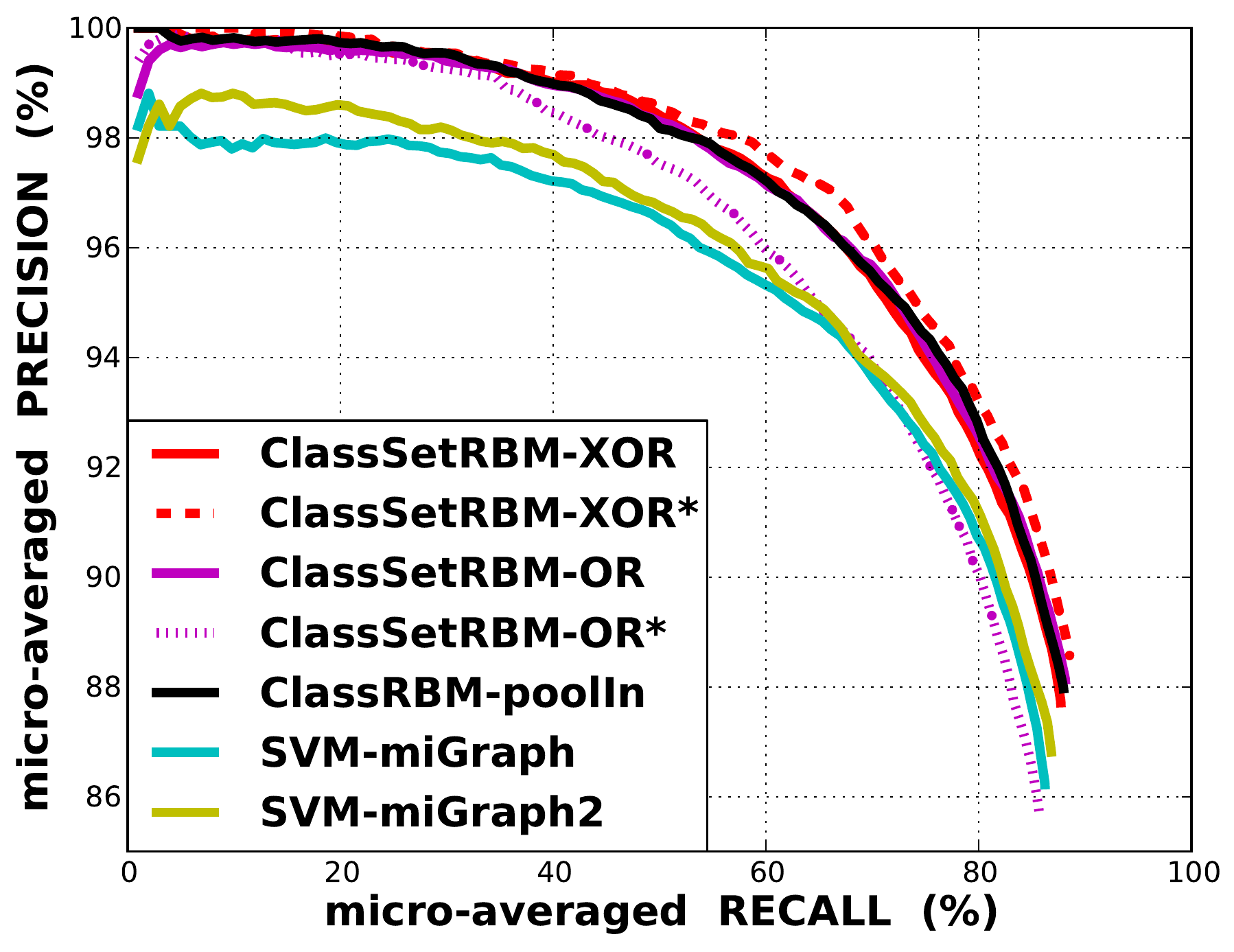}
\centerline{\scriptsize (a) Precision/Recall curves on DsUs}
\end{minipage}
\begin{minipage}{0.49\linewidth}
\includegraphics[width=\linewidth]{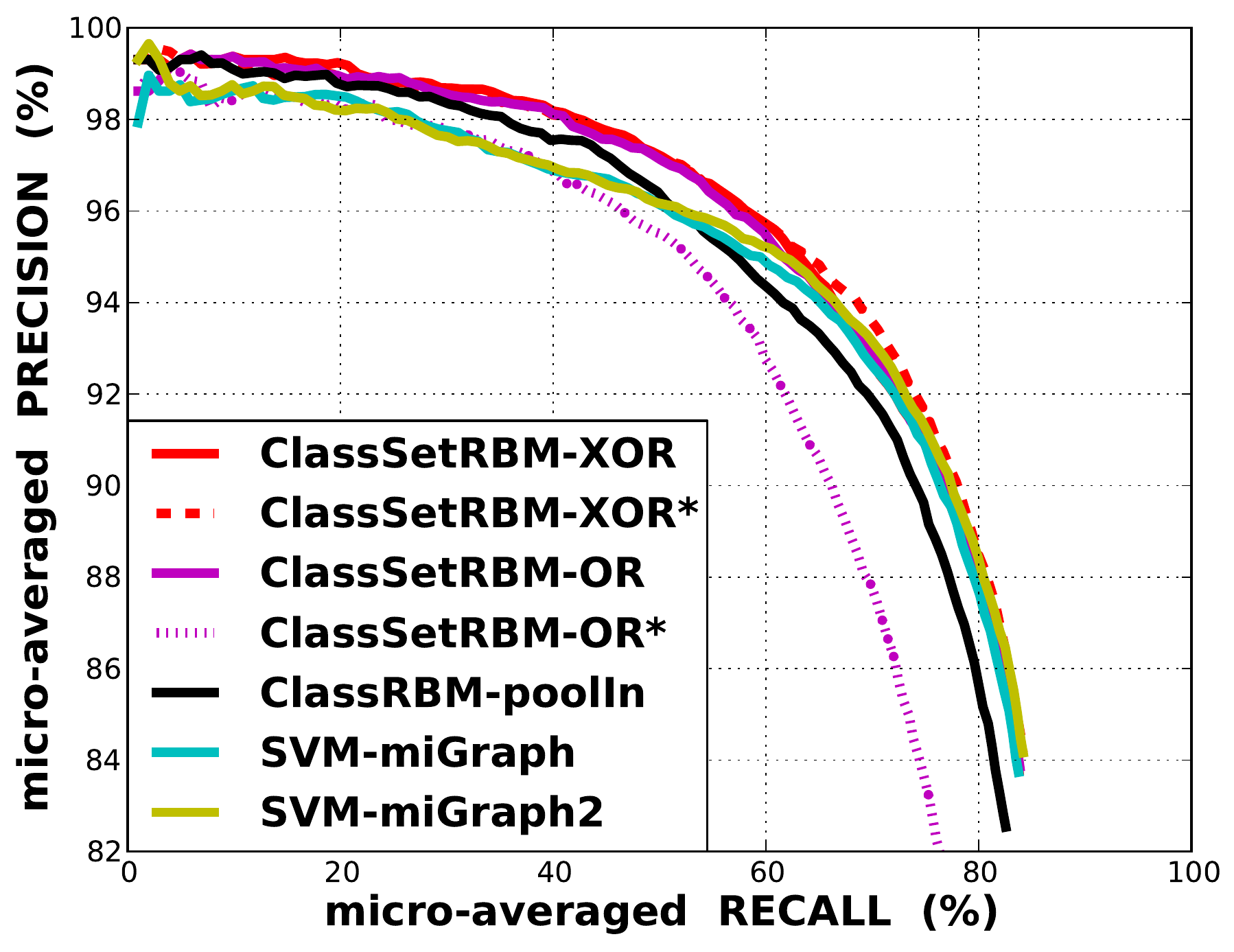}
\centerline{\scriptsize (b) Precision/Recall curves on DsDe}
\end{minipage}
\end{center}
\caption{Precision/recall curves for mail classification problem.}
\label{fig:mail_results}
\end{figure*}

We carried out experiments on two collected datasets for mail classification:
{\it DsDe}
(14\,593 pieces of mail,
102\,071 pages,
11 classes)
and {\it DsUs}
(16\,808 pieces of mail,
160\,372 pages,
6 classes).
Table~\ref{tab:mail_results}(a) shows the 5-fold cross-validation
average accuracy of different models.
It is also important in mailroom automation to be able to reject pieces of
mail for which the prediction is too uncertain: the pieces of mail
rejected by the system can then be processed by a human agent, in
order to limit classification errors of the whole process.  The
rejection is done by comparing the classifier's output 
confidence to a threshold, thus this confidence estimate has to be
reliable.  A standard way to evaluate the goodness of the output
confidence scores in multiclass classification is to plot {\it
  micro-averaged} recall and precision~\citep{Sebastiani2002} for
different values of the rejection threshold.  This is shown in
Figure~\ref{fig:mail_results}(b,c).

\begin{table}[t]
\begin{center}
\begin{tabular}{ l|cc }
\hline
Model                   & DsUs      & DsDe\\
\hline
\proponediscr           &     87.71 & 83.75\\
\proponehardmax         & \BF 88.55 & \BF 84.18 \\
\proptwodiscr           & \bf 88.13 & \bf 83.86 \\
\proptwohardmax         &     85.69 & 78.43     \\
\hline\hline
\inputPoolRBM           & \bf 87.97 & 82.54 \\
\SVMmiGraph             &     86.22 & \bf 83.74 \\
\SVMmiGraphtwo          &     86.82 & \bf 84.16 \\
\SVMlocal               &     56.45 &     64.14 \\
\hline
\end{tabular}
\end{center}
\caption{Classification accuracies (\%) on mail classification task}
\label{tab:mail_results}
\end{table}

Again, we observe that {\proponehardmax} achieves the best performance
overall. We emphasize that, for both these two large datasets,
{\proponehardmax} was much faster to train than the SVM
approaches.

\section{Conclusion}

We described how the classification restricted Boltzmann machine could
be adapted to problems where the inputs correspond to sets of vectors.
Different generalizations for this problem were investigated, with
one of these variant achieving consistent, competitive performance
on multiple-instance learning datasets. It was also applied with success
to a mail categorization task. Our experiments confirm the
usefulness of pooling at the hidden representation level,
as opposed to the input or output level. 
Directions for future work include applying this framework
in deep neural architectures.

\subsection*{Acknowledgements}

Hugo acknowledges the financial support of NSERC\@.

\section*{Appendix}

\subsection*{Derivation of the free-energy for $\propone$}

We provide here the derivation of the free-energy
for $\propone$. To simplify the derivation, 
we assume hidden layer sizes of $H=1$. The generalization
to arbitrary size is trivial, since the necessary sums factorize for each
hidden unit, for the same reason that the conditional
over $\Hbf$ given $\Xbf$ and $\ybf$ factorizes into each of the $j^{\rm th}$ hidden unit sets
$\{h^{s}_j\}$.

\begin{eqnarray}
p( \ybf = \ebf_{\cl} | \xbf ) & = & \sum_{\Hbf} p( \ybf = \ebf_{\cl}, \Hbf | \xbf ) \nonumber\\
        & = & \frac{ \sum_{\Hbf} \exp(-E(\Xbf, \ybf, \Hbf))}{\sum_{\Hbf'}\sum_{\cl'=1\ldots \Cl} \exp(-E(\Xbf, \ebf_{\cl'}, \Hbf'))} \label{eqn:prob_prop_one}
\end{eqnarray}
where
\begin{eqnarray}
 & & \sum_{\Hbf} \exp(-E(\Xbf, \ybf, \Hbf)) \nonumber\\
 & = & \sum_{\Hbf} \exp\left(\dbf^\top \ybf + \sumOverTheBag \bbf^\top \xbf^{(s)} + \sumOverTheBag \cbf^\top \hbf^{(s)} + \sumOverTheBag\Big( {\hbf^{(s)}}^\top \Wbf \xbf^{(s)}  + {\hbf^{(s)}}^\top \Ubf \ybf \Big)\right) \nonumber\\
 & = & \exp(\dbf^\top \ybf + \sumOverTheBag \bbf^\top \xbf^{(s)}) \sum_{\Hbf} \exp\left( \sumOverTheBag \cbf^\top \hbf^{(s)} + \sumOverTheBag\Big( {\hbf^{(s)}}^\top \Wbf \xbf^{(s)}  + {\hbf^{(s)}}^\top \Ubf \ybf \Big)\right) \nonumber\\
 & = & \exp(\dbf^\top \ybf + \sumOverTheBag \bbf^\top \xbf^{(s)}) \left( \exp(0) + \sum_{s} \exp\left( c_1 + \Wbf_{1\cdot} \xbf^{(s)}  + \Ubf_{1\cdot} \ybf \right)\right) \label{eqn:h_mult_excl}\\
 & = & \exp(\dbf^\top \ybf + \sumOverTheBag \bbf^\top \xbf^{(s)}) \left( 1 + \exp(\Ubf_{1\cdot}\ybf ) \sum_{s} \exp\left( c_1 + \Wbf_{1\cdot} \xbf^{(s)} \right)\right) \nonumber
\end{eqnarray}
where at line~\ref{eqn:h_mult_excl} we used the mutual exclusivity constraint $\sum_s h_1^{(s)} \in \{0,1\}$ over $\Hbf$. Hence we can write
\begin{eqnarray*}
\log \sum_{\Hbf} \exp(-E(\Xbf, \ybf, \Hbf)) & = & \dbf^\top \ybf + \sumOverTheBag \bbf^\top \xbf^{(s)} + \Softplus\left( \Ubf_{1\cdot}\ybf + \log\left( \sum_{s} \exp\left( c_1 + \Wbf_{1\cdot} \xbf^{(s)} \right)\right)\right) \\
        & = & \dbf^\top \ybf + \sumOverTheBag \bbf^\top \xbf^{(s)} + \Softplus\left( \Ubf_{1\cdot}\ybf + {\rm softmax}_1(\Xbf) \right) \\
\end{eqnarray*}
where we use the definition of ${\rm softmax}_j(\Xbf)$ for the $\propone$ (see Section~\ref{ssec:prop1}). Going back to Equation~\ref{eqn:prob_prop_one}:
\begin{eqnarray*}
p( \ybf = \ebf_{\cl} | \xbf ) & = & \frac{ \sum_{\Hbf} \exp(-E(\Xbf, \ybf, \Hbf))}{\sum_{\Hbf'}\sum_{\cl'=1\ldots \Cl} \exp(-E(\Xbf, \ebf_{\cl'}, \Hbf'))} \\
    & = & \frac{\exp\left(\dbf^\top \ybf + \sumOverTheBag \bbf^\top \xbf^{(s)} + \Softplus\left( \Ubf_{1\cdot}\ybf + {\rm softmax}_1(\Xbf) \right)\right)}{\sum_{\cl'=1\ldots \Cl} \exp\left(\dbf^\top \ebf_{\cl'} + \sumOverTheBag \bbf^\top \xbf^{(s)} + \Softplus\left( \Ubf_{1\cdot}\ebf_{\cl'}, + {\rm softmax}_1(\Xbf) \right)\right)}\\
    & = & \frac{\exp\left(\dbf^\top \ybf + \Softplus\left( \Ubf_{1\cdot}\ybf + {\rm softmax}_1(\Xbf) \right)\right)}{\sum_{\cl'=1\ldots \Cl} \exp\left(\dbf^\top \ebf_{\cl'} + \Softplus\left( \Ubf_{1\cdot}\ebf_{\cl'}, + {\rm softmax}_1(\Xbf) \right)\right)}\\
    & = & \frac{\exp\left(-F^{\rm XOR}(\Xbf,\ybf)\right)}{\sum_{\cl'=1\ldots \Cl} \exp\left(-F^{\rm XOR}(\Xbf,\ebf_{\cl'})\right)}
\end{eqnarray*}
where we recover $F^{\rm XOR}(\Xbf,\ybf) = -\dbf^\top \ybf - \Softplus\left( \Ubf_{1\cdot}\ybf + {\rm softmax}_1(\Xbf) \right)$ for $H=1$. Because of the hidden unit factorization property,
we get the general free-energy function $F^{\rm XOR}(\Xbf,\ybf) = -\dbf^\top \ybf - \sum_{j=1}^H \Softplus\left( \Ubf_{j\cdot}\ybf + {\rm softmax}_j(\Xbf) \right)$ for arbitrary values of $H$.

\section*{Derivation of the free-energy for $\proptwo$}

Again, we provide the derivation of the free-energy
for $\propone$. Here, we can also simplify the derivation by 
we assuming hidden layers of size $H=1$.
\begin{eqnarray}
p( \ybf = \ebf_{\cl} | \xbf ) & = & \sum_{\Fbf} \sum_{\Hbf} p( \ybf = \ebf_{\cl}, \Hbf, \Fbf | \xbf ) \nonumber\\
        & = & \frac{ \sum_{\Fbf} \sum_{\Hbf} \exp(-E(\Xbf, \ybf, \Hbf, \Fbf))}{\sum_{\Fbf'}\sum_{\Hbf'}\sum_{\cl'=1\ldots \Cl} \exp(-E(\Xbf, \ebf_{\cl'}, \Hbf', \Fbf'))} \label{eqn:prob_prop_two}
\end{eqnarray}
where
\begin{eqnarray}
 & & \sum_{\Fbf} \sum_{\Hbf} \exp(-E(\Xbf, \ybf, \Hbf, \Fbf)) \nonumber\\
 & = & \sum_{\Fbf} \sum_{\Hbf} \exp\left(\dbf^\top \ybf + \sumOverTheBag \bbf^\top \xbf^{(s)} + \sumOverTheBag \cbf^\top \hbf^{(s)} + \sumOverTheBag\Big( {\hbf^{(s)}}^\top \Wbf \xbf^{(s)}  + {\fbf^{(s)}}^\top \Ubf \ybf \Big)\right) \nonumber\\
 & = & \exp(\dbf^\top \ybf + \sumOverTheBag \bbf^\top \xbf^{(s)}) \sum_{\Fbf} \sum_{\Hbf} \exp\left( \sumOverTheBag \cbf^\top \hbf^{(s)} + \sumOverTheBag\Big( {\hbf^{(s)}}^\top \Wbf \xbf^{(s)}  + {\fbf^{(s)}}^\top \Ubf \ybf \Big)\right) \nonumber\\
 & = & \exp(\dbf^\top \ybf + \sumOverTheBag \bbf^\top \xbf^{(s)}) \left( \exp(0) \left(\sum_{\Hbf} \exp\left(  \cbf^\top \hbf^{(s)} + {\hbf^{(s)}}^\top \Wbf \xbf^{(s)}\right)\right) \right. \label{eqn:f_mult_excl}\\
& & ~~~~~~~~~~~~~~~~~~~~~~~~~~~~~~~~~~~~~~~~~~~~~\left.+ \sum_{s} \exp(\Ubf_{1\cdot} \ybf) \left(\sum_{\Hbf ~ {\rm s.t.}~ h_1^{(s)} = 1} \exp\left( \cbf^\top \hbf^{(s)} + \hbf^{(s)} \Wbf \xbf^{(s)} \right)\right)\right) \nonumber\\
 & = & \exp(\dbf^\top \ybf + \sumOverTheBag \bbf^\top \xbf^{(s)}) \left( \exp(0) \prod_{s'}\left(1 + \exp\left(  c_1 + \Wbf_{1\cdot} \xbf^{(s')}\right)\right) \right. \label{eqn:h_f_constraint}\\
& & ~~~~~~~~~~~~~~~~~~~~~~~~~~~~~~~~~~~~~~~~~~~~~\left.+ \sum_{s} \exp(\Ubf_{1\cdot} \ybf + c_1 + \Wbf_{1\cdot} \xbf^{(s)}) \prod_{s'\neq s}\left(1 + \exp\left(  c_1 + \Wbf_{1\cdot} \xbf^{(s')}\right)\right) \right) \nonumber\\
 & = & \exp(\dbf^\top \ybf + \sumOverTheBag \bbf^\top \xbf^{(s)}) \left(\prod_{s'}\left(1 + \exp\left(  c_1 + \Wbf_{1\cdot} \xbf^{(s')}\right)\right) \right) \left( 1 + \exp(\Ubf_{1\cdot} \ybf)\sum_{s} \frac{\exp( c_1 + \Wbf_{1\cdot} \xbf^{(s)})}{1 + \exp( c_1 + \Wbf_{1\cdot} \xbf^{(s)})} \right)\nonumber
\end{eqnarray}
where at line~\ref{eqn:f_mult_excl} we used the mutual exclusivity constraint $\sum_s g_1^{(s)} \in \{0,1\}$ over $\Fbf$, and at line~\ref{eqn:h_f_constraint} we used
the inequality constraint between $\Hbf$ and $\Fbf$,  $h^{(s)}_1 \geq \f^{(s)}_1~\forall~s$. Hence we can write
\begin{eqnarray*}
 & & \log \sum_{\Fbf} \sum_{\Hbf} \exp(-E(\Xbf, \ybf, \Hbf, \Fbf)) \\
& = & \dbf^\top \ybf + \sumOverTheBag \bbf^\top \xbf^{(s)} + \sum_{s'} \Softplus(c_1 + \Wbf_{1\cdot} \xbf^{(s')}) +  \Softplus\left( \Ubf_{1\cdot} \ybf + \log \sum_{s} \frac{\exp( c_1 + \Wbf_{1\cdot} \xbf^{(s)})}{1 + \exp( c_1 + \Wbf_{1\cdot} \xbf^{(s)})} \right)\\
& = & \dbf^\top \ybf + \sumOverTheBag \bbf^\top \xbf^{(s)} + \sum_{s'} \Softplus(c_1 + \Wbf_{1\cdot} \xbf^{(s')}) +  \Softplus\left( \Ubf_{1\cdot} \ybf + \log \sum_{s} \exp( \Softminus(c_1 + \Wbf_{1\cdot} \xbf^{(s)})) \right)\\
& = & \dbf^\top \ybf + \sumOverTheBag \bbf^\top \xbf^{(s)} + \sum_{s'} \Softplus(c_1 + \Wbf_{1\cdot} \xbf^{(s')}) +  \Softplus\left( \Ubf_{1\cdot} \ybf + {\rm softmax}_1(\Xbf) \right)\\
\end{eqnarray*}
where we use the definition of ${\rm softmax}_j(\Xbf)$ for the $\proptwo$ (see Section~\ref{ssec:prop2}). Going back to Equation~\ref{eqn:prob_prop_two}:
\begin{eqnarray*}
p( \ybf = \ebf_{\cl} | \xbf ) & = & \frac{ \sum_{\Hbf} \sum_{\Fbf} \exp(-E(\Xbf, \ybf, \Hbf, \Fbf))}{\sum_{\Hbf'}\sum_{\Fbf'}\sum_{\cl'=1\ldots \Cl} \exp(-E(\Xbf, \ebf_{\cl'}, \Hbf', \Fbf'))} \\
& = & \frac{ \exp\left(\dbf^\top \ybf + \Softplus\left( \Ubf_{1\cdot} \ybf + {\rm softmax}_1(\Xbf) \right)\right)}{\sum_{\cl'=1\ldots \Cl} \exp(\dbf^\top \ebf_{\cl'} + \Softplus\left( \Ubf_{1\cdot} \ebf_{\cl'} + {\rm softmax}_1(\Xbf) \right))} \\
& = & \frac{\exp\left(-F^{\rm OR}(\Xbf,\ybf)\right)}{\sum_{\cl'=1\ldots \Cl} \exp\left(-F^{\rm OR}(\Xbf,\ebf_{\cl'})\right)}
\end{eqnarray*}
where we recover $F^{\rm OR}(\Xbf,\ybf) = -\dbf^\top \ybf - \Softplus\left( \Ubf_{1\cdot} \ybf + {\rm softmax}_1(\Xbf) \right)$ for $H=1$.
Again, because of the hidden unit factorization property of $\proptwo$,
we get the general free-energy function $F^{\rm OR}(\Xbf,\ybf) = -\dbf^\top \ybf - \sum_{j=1}^H \Softplus\left( \Ubf_{j\cdot}\ybf + {\rm softmax}_j(\Xbf) \right)$ for arbitrary values of $H$.

\bibliographystyle{plainnat}
\bibliography{bag_class_rbm_arxiv_bib}
\end{document}